\documentclass[10pt,twocolumn,letterpaper]{article}

\usepackage{iccv}      % To produce the REVIEW version
\usepackage{cuted}

\definecolor{iccvblue}{rgb}{0.21,0.49,0.74}
\usepackage[pagebackref,breaklinks,colorlinks,allcolors=iccvblue]{hyperref}

\usepackage{multirow}
\usepackage{makecell}

\usepackage{mdframed}

\usepackage{listings}
\usepackage{xcolor}

\definecolor{codegreen}{rgb}{0,0.6,0}
\definecolor{codegray}{rgb}{0.5,0.5,0.5}
\definecolor{codepurple}{rgb}{0.58,0,0.82}
\definecolor{backcolour}{rgb}{0.95,0.95,0.92}

\definecolor{backcolour}{rgb}{0.95, 0.95, 0.96}

\lstdefinestyle{mystyle}{
    backgroundcolor=\color{backcolour},   % 设置背景色
    commentstyle=\color{red},
    keywordstyle=\color{blue},
    numberstyle=\tiny\color{gray},
    stringstyle=\color{black},
    basicstyle=\ttfamily\footnotesize,
    breakatwhitespace=false,         
    breaklines=true,                 
    captionpos=b,                    
    keepspaces=true,                 
    numbers=left,                    
    numbersep=5pt,                  
    showspaces=false,                
    showstringspaces=false,
    showtabs=false,                  
    tabsize=2
}

\usepackage{listings}
\usepackage{courier} % 引入 Courier 字体包

\newmdenv[
  backgroundcolor=gray!20,   % 设置背景颜色为灰色
  linecolor=gray!20,         % 边框颜色同背景，可使边框“消失”
  innerleftmargin=10pt,      % 内边距
  innerrightmargin=10pt,
  innertopmargin=10pt,
  innerbottommargin=10pt
]{mytextbox}

\def\paperID{913} % *** Enter the Paper ID here
\def\confName{ICCV}
\def\confYear{2025}

\title{Model as a Game: On Numerical and Spatial Consistency for Generative Games}

\author{%
  Jingye Chen$^{*13}$, Yuzhong Zhao\thanks{Work done during internship at Microsoft Research.}\;$^{23}$, Yupan Huang$^{3}$,  Lei Cui$^{3}$, Li Dong$^{3}$, Tengchao Lv$^{3}$, \\  Qifeng Chen$^{1}$, Furu Wei$^{3}$ \\
  $^{1}$HKUST \;\;\;   $^{2}$UCAS \;\;\; $^{3}$Microsoft Research \\
  \texttt{qwerty.chen@connect.ust.hk,cqf@ust.hk},
  \texttt{\{lecu,fuwei\}@microsoft.com}
}

\begin{document}
\maketitle
% \vspace{-1.7cm}

\begin{strip}
\centering
% \vspace{-1.7cm}
\includegraphics[width=0.98\linewidth]{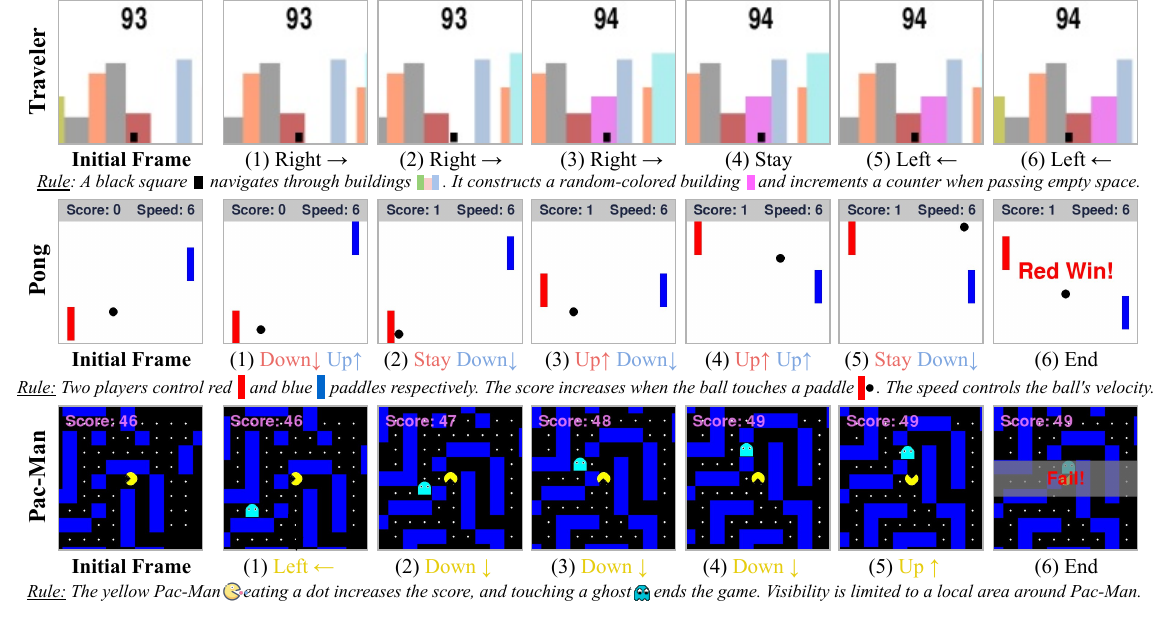}
\captionof{figure}{Demonstration of numerical and spatial consistency in our generative games. We train an action-controllable image generative model for Traveler (top) and extends to other games Pong and Pac-Man (bottom). Action sequences and rules are below each gameplay.\label{fig:teaser}}
\end{strip}

\begin{abstract}
Recent advances in generative models have significantly impacted game generation. However, despite producing high-quality graphics and adequately receiving player input, existing models often fail to maintain fundamental game properties such as numerical and spatial consistency. Numerical consistency ensures gameplay mechanics correctly reflect score changes and other quantitative elements, while spatial consistency prevents jarring scene transitions, providing seamless player experiences.
In this paper, we revisit the paradigm of generative games to explore what truly constitutes a Model as a Game (MaaG) with a well-developed mechanism. We begin with an empirical study on ``Traveler'', a 2D game created by an LLM featuring minimalist rules yet challenging generative models in maintaining consistency.
Based on the DiT architecture, we design two specialized modules: (1) a numerical module that integrates a LogicNet to determine event triggers, with calculations processed externally as conditions for image generation; and (2) a spatial module that maintains a map of explored areas, retrieving location-specific information during generation and linking new observations to ensure continuity.
Experiments across three games demonstrate that our integrated modules significantly enhance performance on consistency metrics compared to baselines, while incurring minimal time overhead during inference.
\end{abstract}

\section{Introduction}
Gaming has become an integral part of modern daily life. Traditional game development adheres to a well-defined pipeline in which storylines and scenes are meticulously crafted in advance, ensuring stable gameplay but limiting player exploration. Recent advances in generative AI have opened new possibilities for game generation using state-of-the-art generative models~\cite{yang2024playable,yu2025gamefactory,che2024gamegen,valevski2024diffusion}. The open-domain capabilities of image, video, and 3D generative models hold immense promise for integration into game creation processes. For example, Oasis \cite{oasis} employs DiT \cite{peebles2023scalable} to directly generate Minecraft scenes and respond to player keyboard inputs.

\begin{figure}[t]
\centering
\includegraphics[width=0.5\textwidth]{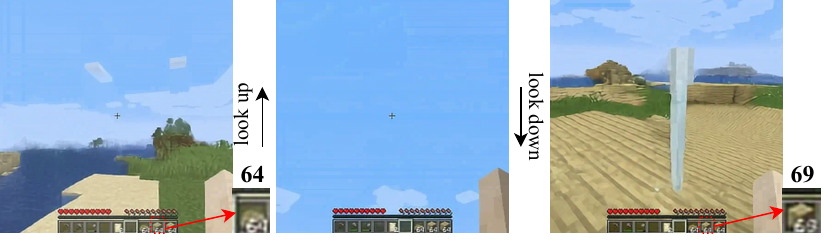}
\caption{While generative games have advanced creative scene generation, significant challenges remain in maintaining numerical and spatial consistency. In Oasis, simply looking up and then down can lead to numerical inconsistencies and spatial discontinuities in the generated environment.}
\vspace{-0.3cm}
\label{intro}
\end{figure}

Current mainstream approaches to generative games predominantly focus on visual generation and simplify game modeling as a pixel prediction problem. However, as illustrated in Figure~\ref{intro}, even though these models offer a degree of playability, they fall short of maintaining crucial aspects of gameplay consistency. We identify two critical forms of consistency for compelling game design: (1) \textit{numerical consistency}, which requires that value changes (e.g., scores, health) correspond accurately to triggering events rather than changing arbitrarily; and (2) \textit{spatial consistency}, which ensures that previously visited areas remain recognizable when revisited.

The challenges of maintaining numerical and spatial consistency are further validated by our empirical study. To investigate these issues in a controlled setting, we develop a minimalist 2D game called Traveler using Pygame (first row of Figure~\ref{fig:teaser}). In Traveler, a black square—the traveler—moves left or right while the camera keeps it centered. The background is populated with colorful buildings that serve as an indicator of map consistency: once established, the building layout should remain unchanged when revisited. In contrast, empty spaces trigger the appearance of new buildings and an update to the score counter. This simple game design decouples the complex tasks of scene rendering from game logic, allowing us to focus specifically on studying consistency in generative gameplay. Our experimental findings on Traveler reveal that existing generative models exhibit significant limitations: they often produce abrupt score fluctuations and fail to consistently reproduce the same background elements, leading to both numerical and spatial discontinuities.
In addition to Traveler, we extend our study to other games such as Pong and Pac-Man (shown in the second and third rows of Figure~\ref{fig:teaser}) to further examine spatial consistency across 1D to 2D maps and numerical consistency across diverse gameplay mechanisms.

To address these challenges, in this paper we advance generative gameplay consistency through two explicitly designed modules. First, we propose a numerical consistency module that integrates a trainable LogicNet to dynamically determine event triggers, paired with an external numerical record that guides the generation process. Second, we design a spatial consistency module that maintains an external map of explored areas, enabling the model to retrieve location-specific data and link new observations back to the original map to preserve spatial coherence. Please note that GameGAN \cite{kim2020learning} also emphasizes the importance of maintaining spatial consistency in games. They uses an external  memory to store the latent map features at each step. Our module not only explicitly records the map but also facilitates the exportation of the player's current location, a critical element in gaming scenarios. It also further simplifies the evaluation process. Moreover, our architecture offers the flexibility to customize maps. Users can preset specific areas for exploration before the game starts or dynamically adjust the map during gameplay. This capability is not possible with designs based solely on latent map features for GameGAN.

Our experimental results validate the effectiveness of the proposed modules. Across three different games, our method successfully mitigates abrupt score fluctuations and ensures that generated observations remain consistent with previously seen data. Quantitative metrics further confirm that our approach significantly outperforms existing baselines in both numerical and spatial consistency. Our key contributions include:

\begin{itemize}
    \item We present the systematic study of numerical and spatial consistency in generative games using specially designed 2D game environments.
    \item We design a novel numerical module that dynamically determines event triggers to enhance numerical consistency in gameplay, and a spatial module that maintains an external map to assist in rendering explored areas, significantly improving spatial consistency.
    \item Experimental validation demonstrates the effectiveness of our consistency modules across three different games.
\end{itemize}

\section{Related Work}

\subsection{Game Creation}

\noindent \textbf{Traditional game creation.} 
The majority of modern video games are developed using game engines such as Unity~\cite{unity} and Unreal Engine (UE)~\cite{unrealengine}. These engines offer powerful tools for designing game mechanics, rendering graphics and simulating physics, which naturally ensures both map and logic consistency of the game. However, game development with these engines is often time-consuming, requiring extensive manual effort in asset creation, game logic programming, and stability testing. Moreover, due to the limited generative capabilities of existing game engines, the games are often constrained by fixed patterns, such as finite worlds and predefined dialogues.

\noindent \textbf{Generative game creation.} In the era of GenAI, game environments can be generated by models, replacing manually designed scenes and enhancing flexibility and imagination \cite{yu2025gamefactory,che2024gamegen,valevski2024diffusion,bruce2024genie,yang2024playable,sudhakaran2023mariogpt,feng2024matrix,genie-2,sudhakaran2023mariogpt,kanervisto2025world,li2024unbounded,ying,menapace2024promptable,menapace2021playable,oasis}.
For instance, GameNGen \cite{valevski2024diffusion} is trained through a two-step process involving a RL-agent learns to play the game for data collection, followed by training a diffusion model. GameGen-X \cite{che2024gamegen} features a two-stage training process that leverages a vast dataset of gameplay clips for pre-training and advanced multi-modal control mechanisms to deliver high-quality, dynamically controllable game content. PGG \cite{yang2024playable} utilizes diffusion forcing \cite{chen2025diffusion}, where it employs memory units similar to RNNs to store historical information, and uses balanced sampling to train a DiT \cite{peebles2023scalable} to generate game scenes.
Muse \cite{kanervisto2025world} enhances creative game development by generating consistent, diverse, and persistent gameplay sequences based on human data. However, despite its claims of ensuring consistency, Muse only considers frame-to-frame continuity while overlooking the broader consistency of the game map. Maintaining visual coherence across distant frames is crucial for a seamless gaming experience, yet it remains a challenge for current models.

Despite significant advancements in game generation as aforementioned, we observe that few works has focused on crucial aspects of games such as numerical and spatial consistency. Although GameGAN \cite{kim2020learning} introduces the use of an external memory module to address spatial consistency, it employs an implicit map representation which introduces several limitations. These include the difficulty in exporting the player's precise location, the lack of support for customizable maps, and challenges in evaluating consistency effectively.

\subsection{Controllable Generative Models}
Generative models have demonstrated significant potential in creating compelling visual outputs \cite{he2024llms,esser2024scaling,kondratyuk2023videopoet,ouyang2024codef,flux}. However, for tailored applications, users often require the integration of conditional controls to steer the generation process towards desired needs. Taking controllable image generation \cite{zhang2023adding,mou2024t2i,zhao2023uni,huang2023composer,peng2024controlnext,chen2023textdiffuser,chen2024textdiffuser,li2024controlnet++} as an example, ControlNet \cite{zhang2023adding} replicates the backbone architecture of Stable Diffusion \cite{rombach2022high} to incorporate additional control conditions such as Canny edge maps and segmentation masks, assist in guiding the image generation. For controllable video generation \cite{chen2023control,hu2023videocontrolnet,he2024cameractrl,yang2024direct,zhang2023controlvideo,zhou2024trackgo,zhang2024interactivevideo,zhang2024world}, CameraCtrl \cite{he2024cameractrl} adds a plug-and-play module to text-to-video models, allowing for accurate control of camera movement and trajectory, thereby improving video narrative depth and customization.

Inspired by these works, we explore the use of explicit numerical and spatial guidance to the game generation.

\begin{figure*}[t]
\centering
\includegraphics[width=1\textwidth]{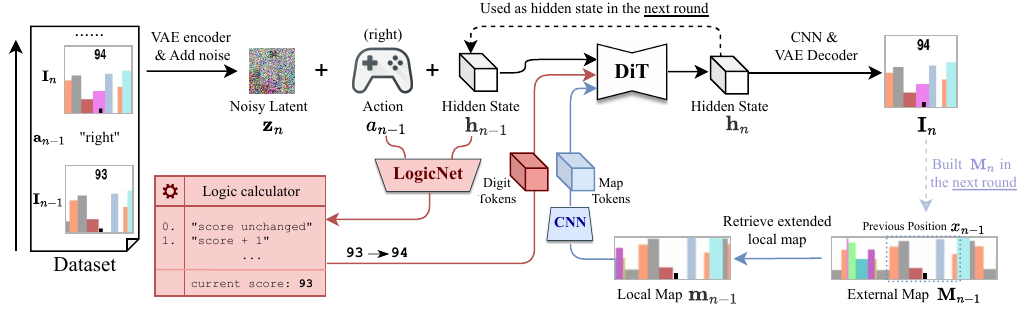}
\caption{
Overall architecture for enhancing the consistency of generative games. The \textbf{black} arrow and components signify the baseline architecture, while the {\color{red}{\textbf{red}}} and {\color{blue}{\textbf{blue}}} arrows and components represent our proposed numerical and spatial modules, respectively. The numerical module utilizes a learnable LogicNet to determine the occurrence of events, and the values calculated by an external logic calculator are then used as conditions for the Diffusion Transformer (DiT) generation. The spatial module maintains an external map that is used to retrieve an extended local map, serving as auxiliary information for generation. New observations are linked to the map for subsequent frames. For example, since a new pink building is added, it is updated on the map accordingly. Note that in the visualization, the upper score part is omitted, while the remaining parts are used to perform matching.}
% \vspace{-0.2cm}
\label{fig:architecture}
\end{figure*}

\section{Methodology}

We present a framework to enhance both numerical and spatial consistency in generative games. Our approach addresses the limitations of existing methods that treat game generation primarily as a pixel prediction problem. After introducing a minimalist game called Traveler and specifying the baseline architecture in Section~\ref{sec:preliminary}, we enhance this baseline in two key aspects: numerical consistency through a LogicNet (Section~\ref{sec:logic}) and spatial consistency through an external map (Section~\ref{sec:map}). We then demonstrate how these proposed modules can be adapted to train various types of generative games in Section~\ref{sec:other_games}.

\subsection{Preliminary}
\label{sec:preliminary}
\noindent \textbf{Detailed game rules of Traveler.} As shown in the first row of Figure~\ref{fig:teaser}, we develop the game Traveler using Pygame to investigate numerical and spatial consistency of generative games in a minimalist way. The game is governed by the following key rules:

\begin{itemize}
    \item The ``black square'', positioned at the center of the screen, acts as a free traveler. It can move left, move right, or remain stationary. The camera continuously follows the traveler, ensuring it stays centered.
    \item The ``colorful buildings'', filling the background, serve as indicators of map consistency. As the traveler moves back and forth, the buildings it encounters must remain unchanged from their previous state, maintaining a seamless and coherent game environment.
    \item The ``empty spaces'', scattered among the colorful buildings, serve as interaction points. When the traveler passes through these empty spaces, new buildings of random colors appear, and the counter increases by one each time. Upon returning to the same location, the traveler should see the newly constructed building rather than an empty space, as the map should be updated.
    \item The ``score above frames'', e.g., ``93/94'' in Figure~\ref{fig:teaser}, records the number of times the traveler passes through empty spaces.
\end{itemize}

\noindent \textbf{Baseline architecture.}
We build upon the architecture from PGG \cite{yang2024playable} that employs an RNN-like diffusion framework, as shown by the ``black arrows'' in Figure \ref{fig:architecture}. During each training step, $N$ consecutive frame and action pairs $\{\mathbf{I}_{n}, a_n\}_{n=1}^N$ are sampled from the dataset and processed through $N$ sequential training iterations, where the hidden state is propagated throughout the sequence.
% Specifically, for each frame $\mathbf{I}_{n} \in \sR^{H\times W\times 3}$, a pre-trained VAE \cite{kingma2013auto} compresses it into a latent feature, which is combined with Gaussian noise to obtain the noisy latent $\mathbf{z}_{n}$ following DDPM \cite{ho2020denoising}, $\mathbf{z}_{n}\in \sR^{\frac{H}{4} \times \frac{W}{4} \times 4}$. 
For each frame $\mathbf{I}_{n} \in \mathbb{R}^{H\times W\times 3}$, a pre-trained Variational Autoencoder (VAE) \cite{kingma2013auto} compresses it into a latent representation, which is combined with Gaussian noise to obtain the noisy latent $\mathbf{z}_{n} \in \mathbb{R}^{\frac{H}{4} \times \frac{W}{4} \times 4}$ following the Denoising Diffusion Probabilistic Model (DDPM) \cite{ho2020denoising}.

The noisy latent $\mathbf{z}_{n}$, the action $a_{n-1}$, and the hidden state of the previous time step $\mathbf{h}_{n-1} \in \mathbb{R}^{\frac{H}{4} \times \frac{W}{4} \times 32}$ are then fed into a Diffusion Transformer (DiT) \cite{peebles2023scalable} for generation. Specifically, $\mathbf{z}_{n}$ and $\mathbf{h}_{n-1}$ are concatenated along the channel dimension before being processed by DiT, while $a_{n-1}$ is integrated via cross-attention. At the first time step, the hidden state is initialized with a learnable embedding.

The DiT outputs an updated hidden state $\mathbf{h}_n$, which is then fed into a lightweight CNN to predict the added Gaussian noise. The predicted noise is used to reconstruct the input frame through the VAE decoder. The denoising objective function is defined as

\begin{align}
    \mathcal{L}_{denoise} & = \mathbb{E}_{\epsilon, n}\Big[||\epsilon - \epsilon_{\theta,n}(\mathbf{z}_n, a_{n-1}, \mathbf{h}_{n-1})||_2^2\Big], \\ 
    \mathbf{z}_n & = \sqrt{\overline{\alpha}_{t}} \, \texttt{VAE}(\mathbf{I}_{n}) + \sqrt{1 - \overline{\alpha}_{t}} \, \epsilon, \label{eq:zt}
\end{align}
where $\epsilon$ is the sampled Gaussian noise, $\epsilon_{\theta,n}$ represents the DiT model parameterized by $\theta$ and conditioned on the action $a_{n-1}$ and the hidden state $\mathbf{h}_{n-1}$, and $\overline{\alpha}_{t}$ is a weighting factor that influences the degree of noise added for a sampled noise level $t$, as introduced in DDPM \cite{ho2020denoising}.

This baseline architecture treats game generation primarily as a pixel prediction problem, which fails to adequately capture numerical and spatial consistency. Our proposed improvements address these limitations directly.

\subsection{Numerical Module}
\label{sec:logic}
As depicted by the ``red arrows and components'' in Figure \ref{fig:architecture}, we propose a trainable network called LogicNet to enhance numerical consistency. LogicNet takes the previous hidden state $\mathbf{h}_{n-1}$ and action ${a}_{n-1}$ as inputs to predict whether logic events should be triggered in the current frame. By precomputing these events and incorporating them as conditions in the diffusion process, we significantly improve numerical consistency.
Specifically, LogicNet first encodes $\mathbf{h}_{n-1}$ into features using two convolutional and downsampling layers. These features, combined with a learnable action embedding, are processed through several MLP layers to predict whether an event will be triggered. The detailed architecture is provided in the Appendix A. In the Traveler game, when the traveler moves over an empty space, LogicNet should correctly predict that a ``score +1'' event will occur. LogicNet is trained to perform binary classification using a cross-entropy loss $\mathcal{L}_{logic}$, and the overall objective function is defined as:

\begin{equation}
    \mathcal{L} = \mathcal{L}_{denoise} + \lambda \mathcal{L}_{logic}, \label{eq:overall}
\end{equation}
where $\lambda$ is a hyperparameter that balances the two loss terms.

Additionally, we maintain an external numerical record to handle logic computations explicitly. As shown in Figure \ref{fig:architecture}, the operation of adding `1' to `93' produces the current score of `94'. To ensure accurate rendering of these values, we follow TextDiffuser-2 \cite{chen2024textdiffuser} by decomposing the score at the character level into the hundreds, tens, and units digits — `0', `9', and `4' respectively. These are integrated into the DiT as \textit{digit tokens} using learnable numerical embeddings. This explicit approach alleviates the numerical computation burden on the DiT, allowing it to focus primarily on visual rendering.

\begin{figure}[t]
\centering
\includegraphics[width=0.48\textwidth]{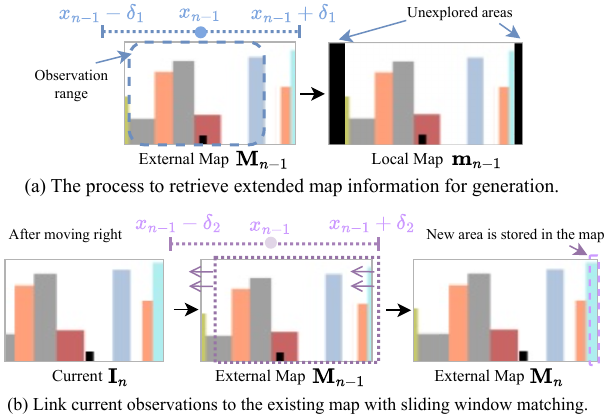}
\caption{Details of the spatial module for retrieval and linking.}
\vspace{-0.3cm}
\label{fig:map}
\end{figure}

\subsection{Spatial Module}
\label{sec:map}
As illustrated by the ``blue arrows and components'' in Figure \ref{fig:architecture}, our spatial module addresses the limitations of RNN-like diffusion models in maintaining spatial consistency. While these models use hidden states to carry temporal information, they often struggle with preserving spatial consistency across frames.
Our approach explicitly stores and updates a persistent map $\mathbf{M}$ that records all previously encountered areas. Each time the player explores a new area, it is integrated into $\mathbf{M}$ while simultaneously updating the player's current location. When generating the next $n\text{th}$ frame $\mathbf{I}_{n}$, we retrieve the current local map $\mathbf{m}_{n-1}$, which is larger than the observed area, to provide auxiliary generative signals for regions outside the current view, ensuring spatial consistency.

For the Traveler game, which is a side-scrolling game where the black square moves horizontally, we structure the map as a horizontally infinite representation. The spatial module performs two key functions:

(1) \textit{Retrieve auxiliary information from the external map for controllable generation} As shown in Figure \ref{fig:map}(a), given a stored map $\mathbf{M}_{n-1}$ and the traveler's central position $x_{n-1}$, we extract a local map $\mathbf{m}_{n-1}$ covering the range $(x_{n-1} - \delta_1, x_{n-1} + \delta_1)$. The parameter $\delta_1$ is set such that $2\delta_1 > W$, providing context beyond the current observation width $W$. Unobserved regions are represented as black pixels. The local map is processed through several CNN layers to generate \textit{map tokens}, which are fed to the DiT as conditional tokens for generation. The detail of the CNN is in Appendix B.

(2) \textit{Link new observations to the external map.} As depicted in Figure \ref{fig:map}(b), we employ a sliding window matching approach to integrate newly observed areas into the map. When a new frame $\mathbf{I}_n$ is generated, we slide it across the existing map within the range $(x_{n-1} - \delta_2, x_{n-1} + \delta_2)$, where $\delta_2$ is determined by the movement scale of the game. For each position, we compute the PSNR for overlapping areas, disregarding regions that extend beyond the existing map. This process determines the new player location $x_n$ and ensures that new observations align seamlessly with previously generated content, preserving spatial consistency throughout gameplay. We showcase the effectiveness of the matching process in Appendix C.

\subsection{Module Generalization to Diverse Games}
\label{sec:other_games}
Our proposed modules are flexible and can be adapted to train various types of games with minor modifications:

\begin{itemize}
    \item For the numerical module, LogicNet's simple design accommodates different game mechanics. In Pong, the critical event is the ball hitting the paddle, while in Pac-Man, it's eating the white dots. For games with multiple simultaneous events, LogicNet can be extended to perform multi-class classification or equipped with multiple prediction heads for combined classification and regression tasks.
  
    \item The spatial module can be extended to 2D maps, as demonstrated in our Pac-Man implementation. The sliding window matching algorithm works along both horizontal and vertical axes, allowing the local map to expand in two dimensions (see Appendix D). For games like Pong that don't require persistent maps, this module can be omitted.
    \item Additional game-specific information, such as ball speed and action embeddings for other characters in multiplayer games such as Pong, can be incorporated through cross-attention mechanisms in the DiT.
\end{itemize}

\section{Experiments}

\begin{figure*}[t]
\centering
\includegraphics[width=1\textwidth]{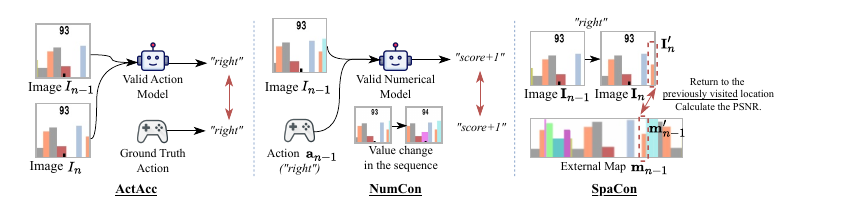}
\caption{Illustration of our three evaluation metrics: Action Accuracy (ActAcc), Numerical Consistency (NumCon), and Spatial Consistency (SpaCon). For ActAcc and NumCon, additional validation models are trained for assessment. For SpaCon, PSNR is calculated by comparing current observations with previous ones at the explored locations.}
\vspace{-0.3cm}
\label{fig:metrics}
\end{figure*}

In this section, we evaluate our approach on three different games, demonstrating the effectiveness of our proposed consistency modules.

\subsection{Implementation Details}

\noindent \textbf{Dataset.} We develop three games (Traveler, Pong, and Pac-Man) using the Pygame library to train and evaluate our proposed modules. The games are created with assistance from Claude 3.5 \cite{claude}, with details provided in the Appendix E. All games run at 30 FPS, with 5\% of the collected data reserved for evaluation. The collection for each game is as follows:
\begin{itemize}
\item \textbf{Traveler:} We generate 100K episodes, each containing 48 frames at $96 \times 96$ resolution. The traveler (a black square) performs random actions selected from ``left'', ``right'', or ``stay''.
\item \textbf{Pong:} We generate 30K episodes with variable frame counts at $128 \times 128$ resolution. Rule-based AI agents control the left and right paddles, executing actions such as ``up'', ``down'', or ``stay'', with randomness incorporated. After game-ending events, we record an additional 30 frames.
\item \textbf{Pac-Man:} We generate 30K episodes with variable frame counts at $128 \times 128$ resolution. Pac-Man randomly selects actions from ``up'', ``down'', ``left'', ``right'', or ``stay'' until encountering a ``monster'', which ends the episode. An additional 30 frames are recorded after each encounter.
\end{itemize}
In addition to video frames, we record supplementary information such as actions and scores. For games requiring spatial awareness (Traveler and Pac-Man), we pre-construct local maps offline for training using a force teaching approach. During inference, maps are constructed progressively based on observed gameplay. The map extension parameters $(\delta_{1},\delta_{2})$ are set to (64, 10) for Traveler and (74, 10) for Pac-Man.

\noindent \textbf{Training.} We adopt the architecture of Playable Game Generator (PGG), which features a lightweight design to minimize inference latency. More details of the architecture are in Appendix F. We utilize their released Variational Autoencoder (VAE) as a foundation for training on our datasets. The trainable components of our generation model include a 33M DiT (Diffusion Transformer), a 0.6M LogicNet in the numerical module, and a 0.1M CNN in the spatial module. Notably, our proposed consistency modules add less than 2\% to the total parameter count of the generation model.

The generation model is trained for 6 epochs on 8 $\times$ 40G A100 GPUs with a batch size of 64, taking approximately 3 days to complete. Since the LogicNet converges quickly, we assign it a lower weight of $\lambda = 1e-4$. For each training iteration, we randomly select an episode and sample a sequence of 32 consecutive frames. The VAE is trained separately for two epochs for each game, requiring one day. For all games, we use three digit tokens as additional input to the DiT. Depending on the input size, we append 40 map tokens for Traveler and 48 for Pac-Man. We set the learning rate to $3e-4$ throughout training.

\subsection{Evaluation Metrics}

To comprehensively evaluate our approach, we employ metrics covering four aspects, illustrations of the last three metrics are shown in Figure \ref{fig:metrics}.

\begin{itemize}
    \item \textbf{Generation quality} measures the visual fidelity and stability of generated content using \underline{FID}~\cite{fid} and \underline{FVD}~\cite{fvd}. FID evaluates individual frames, while FVD assesses temporal coherence in generated video sequences. Note that traditional metrics like PSNR and LPIPS~\cite{zhang2018unreasonable} are unsuitable in our context since new scenes beyond the observed data are randomly generated during inference, making ground truth comparisons impossible.
    
    \item \textbf{Action accuracy} (\underline{ActAcc})~\cite{yang2024playable} evaluates whether predicted actions correctly reflect the actual inputs. Following PGG~\cite{yang2024playable}, we train a Valid Action Model to predict actions between consecutive frames and calculate the percentage of correctly predicted actions.
    
    \item \textbf{Numerical consistency} (\underline{NumCon}) is our newly proposed metric that assesses whether numerical attributes ($e.g.,$ scores) change correctly during gameplay. We train a Valid Numerical Model to predict numerical changes between consecutive frames and calculate an F-measure to evaluate accuracy. For baselines without explicit numerical score tracking, we employ TrOCR \cite{li2023trocr} to recognize numerical values within images.
    
    \item \textbf{Spatial consistency} (\underline{SpaCon}) is another metric we propose to evaluate whether newly revealed areas remain consistent with previously observed locations. We define this metric as:
    $\text{SpaCon} := \mathbb{E}[\text{PSNR}(\textbf{I}^{\prime}_{n}, \textbf{m}{^\prime}_{n-1})]$,
    where $\textbf{I}^{\prime}_{n}$ represents newly visible areas at time step $n$, and $\textbf{m}{^\prime}_{n-1}$ denotes the corresponding map records from time step $n-1$. Please note that GameGAN \cite{kim2020learning} relies solely on comparing the amount specific pixel colors in observations to determine if they match previous observations, a method we believe is prone to significant errors. In contrast, our spatial module enables the precise determination of the player's location. Based on this, the metrics we have designed are inherently more rational and reliable.

\end{itemize}

We provide the details of the additional trained validation models in Appendix G.

\begin{table*}[t]
    \centering
    \footnotesize
    \renewcommand{\arraystretch}{1.2}
    \setlength{\tabcolsep}{2pt}
    \caption{Quantitative results for the three games to investigate the impact of consistency modules. ``D. / P.'' denotes ``Denoising Steps / Prediction Lengths''. Overall, the Consistency Modules significantly boost the baseline in terms of numerical and spatial consistency.}

\begin{tabular}{ccccccccccccccccc}
\toprule
\multirow{2}{*}{\#} & \multirow{2}{*}{D. / P.} & \multirow{2}{*}{\makecell{Consistency \\ Modules}} & \multicolumn{5}{c}{\textbf{Traveler}} & \multicolumn{4}{c}{\textbf{Pong}} & \multicolumn{5}{c}{\textbf{Pac-Man}}\\
\cmidrule(lr){4-8} \cmidrule(lr){9-12} \cmidrule(lr){13-17}
 &  &  & FID$\downarrow$ & FVD$\downarrow$ & ActAcc$\uparrow$ & NumCon$\uparrow$ & SpaCon$\uparrow$ & FID$\downarrow$ & FVD$\downarrow$ & ActAcc$\uparrow$ & NumCon$\uparrow$ & FID$\downarrow$ & FVD$\downarrow$ & ActAcc$\uparrow$ & NumCon$\uparrow$ & SpaCon$\uparrow$\\
\midrule
1 & 8/256 &  & 56.85 & 84.93 & 0.9657 & 0.3245 & 16.15 & 29.30 & \textbf{73.62} & 0.6534 & 0.0889 & 21.70 & 1472.87 & \textbf{0.7382} & 0.2667 & 13.54\\
2 & 8/256 & \checkmark & \textbf{43.76} & \textbf{51.58} & \textbf{0.9909} & \textbf{0.9141} & \textbf{33.64} & \textbf{24.67} & 78.50 & \textbf{0.7871} & \textbf{0.6847} & \textbf{15.43} & \textbf{793.96} & 0.6862 & \textbf{0.6087} & \textbf{18.56} \\
\midrule
3 & 16/256 &  & 51.62 & 83.95 & 0.9658 & 0.3252 & 16.01 & 26.40 & \textbf{78.22} & 0.6161 & 0.0465 & 19.61 & 1194.74 & 0.7260 & 0.3871 & 13.70\\
4 & 16/256 & \checkmark & \textbf{43.75} & \textbf{52.12} & \textbf{0.9916} & \textbf{0.9219} & \textbf{31.39} & \textbf{25.01} & 82.18 & \textbf{0.8717} & \textbf{0.5911} & \textbf{14.24} & \textbf{751.17} & \textbf{0.7869} & \textbf{0.7917} & \textbf{17.88} \\
\bottomrule
\end{tabular}

    \label{tab:metrics}
\end{table*}

\begin{table*}[t]
    \centering
    \footnotesize
    \renewcommand{\arraystretch}{1.2}
    \setlength{\tabcolsep}{2pt}
    \caption{Quantitative results for the three games under different Denoising Steps / Prediction Lengths.}

\begin{tabular}{ccccccccccccccccc}
\toprule
\multirow{2}{*}{\#} & \multirow{2}{*}{D. / P.} & \multicolumn{5}{c}{\textbf{Traveler}} & \multicolumn{4}{c}{\textbf{Pong}} & \multicolumn{5}{c}{\textbf{Pac-Man}}\\
\cmidrule(lr){3-7} \cmidrule(lr){8-11} \cmidrule(lr){12-16}
 &  & FID$\downarrow$ & FVD$\downarrow$ & ActAcc$\uparrow$ & NumCon$\uparrow$ & SpaCon$\uparrow$ & FID$\downarrow$ & FVD$\downarrow$ & ActAcc$\uparrow$ & NumCon$\uparrow$ & FID$\downarrow$ & FVD$\downarrow$ & ActAcc$\uparrow$ & NumCon$\uparrow$ & SpaCon$\uparrow$\\
\midrule
1 & 8/64 & 47.43 & 79.06 & 0.9894 & \textbf{0.9315} & 33.01 & 26.67 & 129.19 & \textbf{0.8622} & 0.6286 & 20.75 & 995.73 & \textbf{0.7643} & 0.5263 & \textbf{20.85}\\
2 & 8/128 & 45.69 & 61.92 & 0.9900 & 0.9054 & \textbf{34.88} & 24.97 & 98.51 & 0.8044 & \textbf{0.7593} & 17.20 & \textbf{789.25} & 0.7256 & 0.6667 & 18.80\\
3 & 8/256 & \textbf{43.76} & \textbf{51.58} & \textbf{0.9909} & 0.9141 & 33.64 & \textbf{24.67} & \textbf{78.50} & 0.7871 & 0.6847 & \textbf{15.43} & 793.96 & 0.6862 & \textbf{0.6087} & 18.56 \\
\midrule
4 & 16/64 & 45.65 & 75.63 & 0.9902 & \textbf{0.9493} & \textbf{34.04} & 28.68 & 125.39 & 0.8613 & 0.6415 & 20.72 & 835.93 & \textbf{0.8510} & 0.6667 & \textbf{18.31}\\
5 & 16/128 & 44.04 & 53.12 & 0.9882 & 0.9323 & 31.55 & 27.11 & 83.17 & 0.8664 & \textbf{0.6667} & 16.74 & 774.92 & 0.8343 & 0.7586 & 18.29\\
6 & 16/256  & \textbf{43.75} & \textbf{52.12} & \textbf{0.9916} & 0.9219 & 31.39 & \textbf{25.01} & \textbf{82.18} & \textbf{0.8717} & 0.5911 & \textbf{14.24} & \textbf{751.17} & 0.7869 & \textbf{0.7917} & 17.88 \\
\bottomrule
\end{tabular}
    \label{tab:ablation_dp}
\end{table*}

\subsection{Quantitative Results}
Tables~\ref{tab:metrics} and~\ref{tab:ablation_dp} present our comprehensive quantitative evaluation. For FID calculation, we randomly sample 5,000 images from both the predicted results and ground truth. For all other metrics, we generate 100 episodes for evaluation.

\noindent \textbf{Effectiveness of the proposed consistency modules.} As shown in Table~\ref{tab:metrics}, incorporating our consistency modules (numerical and spatial) significantly improves performance compared to the baseline. For the Traveler game (rows 1-2), our modules boost numerical consistency (NumCon) by 0.5896 (from 0.3245 to 0.9141) and spatial consistency (SpaCon) by 17.49 (from 16.15 to 33.64). These substantial improvements demonstrate the effectiveness of our numerical module for logical calculations and spatial module for map maintenance. Similar improvements are observed across other games, confirming that our consistency modules generalize well to different applications.

\noindent \textbf{Influence of denoising steps.} Table~\ref{tab:ablation_dp} reveals that increasing the number of denoising steps from 8 to 16 does not always improve performance. For the Pong game, we observe decreased performance in FID and FVD metrics with more denoising steps, suggesting that additional computational cost does not guarantee better results. This observation aligns with findings reported in PGG~\cite{yang2024playable}.

\noindent \textbf{Influence of prediction lengths.} As shown in Table~\ref{tab:ablation_dp}, extending the sequence length from 64 to 256 frames does not degrade generation quality. This important finding demonstrates that our approach can iteratively generate content for indefinite gameplay while maintaining consistent quality.

\begin{figure*}[t]
\centering
\includegraphics[width=1\textwidth]{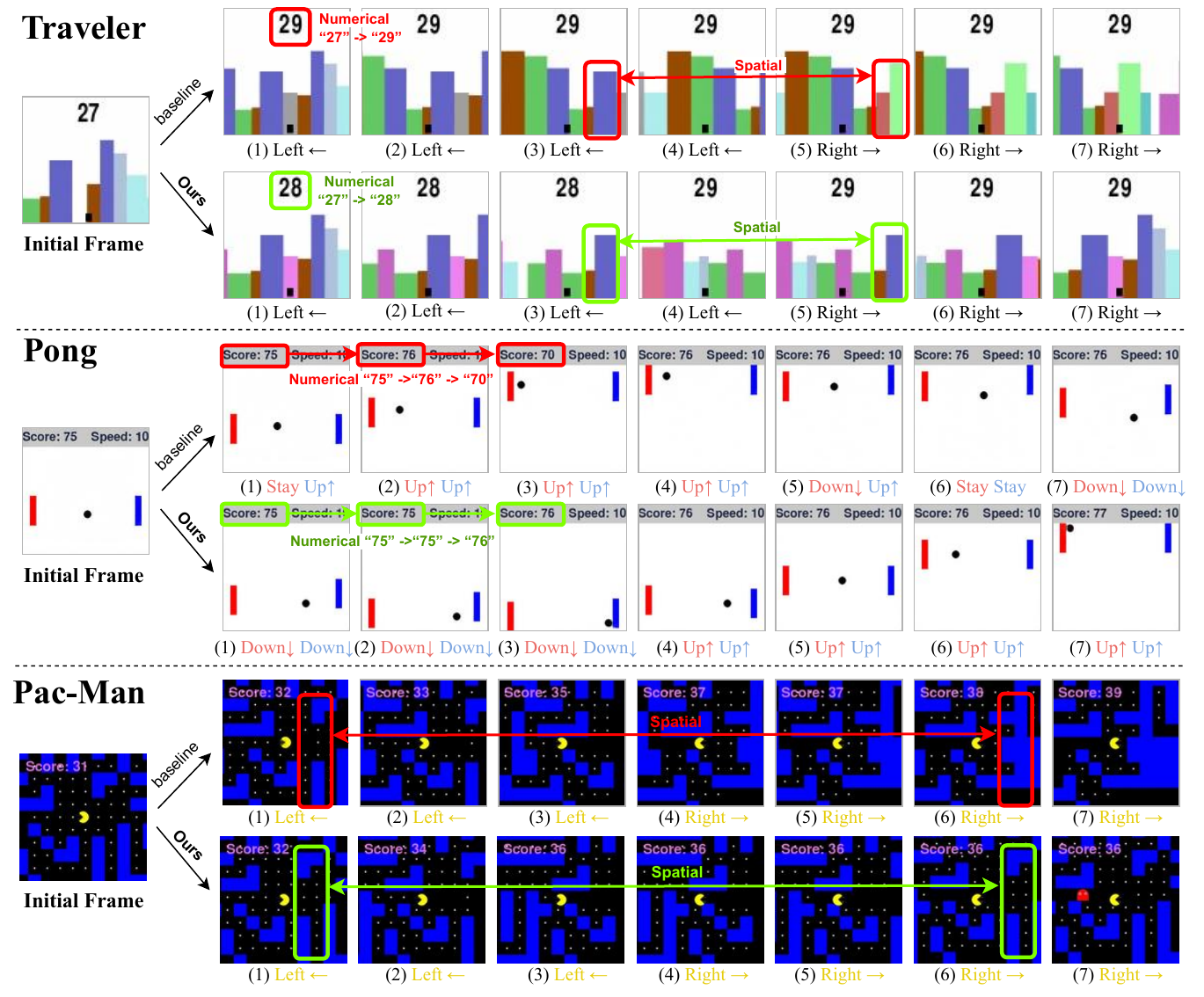}
\caption{Qualitative comparisons between our method and the baseline. We highlight consistent and inconsistent parts using green and red dashed boxes. The consistency problems is divided into two types, which are marked as ``Numerical'' and ``Spatial'' in the figure respectively. Equipped with the proposed consistency modules, our methods achieve better visual performance compared to the baseline.}
\vspace{-0.3cm}
\label{fig:qual}
\end{figure*}

\subsection{Qualitative Results}
Figure~\ref{fig:qual} presents qualitative comparisons between our method and the baseline across all three games. The baseline exhibits clear deficiencies in maintaining both numerical and spatial consistency. 

Regarding numerical consistency, in Traveler, the baseline erroneously increases the score from '27' to '29' after the black square passes a single empty space—an inconsistency that our method avoids. In Pong, the baseline's score inexplicably fluctuates from '75' to '76' to '70', severely compromising the gaming experience. Our method maintains logical numerical progression throughout.

For spatial consistency, our approach ensures that previously encountered map areas remain unchanged when revisited during navigation. In contrast, the baseline regenerates entirely new maps, failing to maintain spatial coherence with previously observed areas. This difference is particularly noticeable in back-and-forth movement scenarios.

Overall, our qualitative results confirm that our method produces more consistent and believable gameplay experiences. The integrated consistency modules successfully address the limitations of existing approaches, enabling coherent game generation with stable numerical attributes and spatial layouts.

\section{Discussion}

\noindent \textbf{Computational efficiency despite architectural complexity.} A significant challenge in enhancing game generation models is maintaining performance while adding consistency features. Our experiments demonstrate that we achieve this balance effectively. We establish a website demo running on a single A100 40G GPU. Despite our additional consistency modules, the frames per second (FPS) remain almost identical to the baseline. When using 8 sampling steps, all three games maintain 10 FPS, decreasing to 6 FPS with 16 sampling steps. The LogicNet requires only 0.0004 seconds during inference, while map retrieval and linking takes 0.015 seconds, with a total memory of just 1.2 GB. This performance efficiency is crucial for interactive games, as it enables responsive gameplay while ensuring logical consistency. Attempts to further accelerate generation by reducing sampling steps to 4 resulted in unstable outputs, suggesting our current configuration represents an optimal balance between speed and generation quality.

\begin{figure}[t]
\centering
\includegraphics[width=0.5\textwidth]{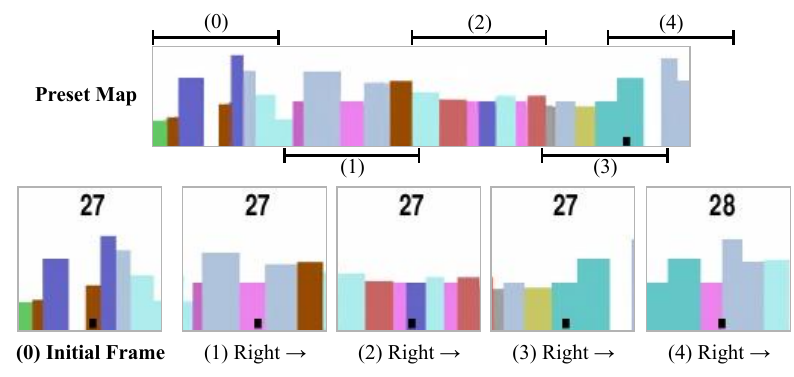 }
\caption{The ability to support map customization. The below part is the inference observation sequence, which exactly corresponds to the pre-set map.}
\vspace{-0.3cm}
\label{fig:map_customization}
\end{figure}

\noindent \textbf{The ability to support map customization.} As demonstrated in Figure \ref{fig:map_customization}, we have pre-set the map in the traveler game. During gameplay, when the \textit{moving right} actions are taken, the areas observed correspond exactly to those pre-defined on the map. This level of controllability is crucial if game designers wish to incorporate specific elements and ideas into the game development process. It is important to note that GameGAN \cite{kim2020learning} cannot achieve this level of control because it does not maintain an explicit map, but rather a latent map which remains inaccessible and unmodifiable by game designers.

\noindent \textbf{Accuracy of text rendering guided by digit tokens.} We employ TrOCR \cite{li2023trocr} to recognize scores in the generated images and observe that in 99\% of cases, these scores align with the intended digit token conditions, demonstrating the effectiveness of our conditional approach.

\noindent \textbf{User study.} We conducted a user study with eight participants to evaluate the gameplay experience. All participants reported that our consistency-enhanced version demonstrated superior playability compared to the baseline. Participants also provided constructive feedback, suggesting that increasing the frame rate to at least 30 FPS would improve the overall experience.

\noindent \textbf{Failure cases.} The sliding window matching technique can fail when encountering continuously repetitive scenes, such as backgrounds consisting solely of similarly-colored buildings. Visualizations are provided in the Appendix H.

\section{Conclusion}

In this paper, we present a novel framework addressing a critical yet under-explored challenge in generative game creation: maintaining consistency throughout the game experience. We focus on two fundamental aspects, including numerical and spatial consistency, which are essential for creating coherent and playable generative games. Experiments across three  games demonstrate significant improvements in both aspects of consistency. For \textit{future work}, we plan to extend our framework to more complex and dynamic 2D and 3D environments and develop more robust spatial consistency modules. This work establishes a foundation for consistency-aware generative game creation, advancing the development toward more coherent gaming experiences.
\onecolumn
\appendix

\noindent {\Large{\textbf{Appendix}}}
% \section{Introduction of the Attached Zip File}

% The attached zip file includes training code, inference code, model, and dataset creation code. We also provide samples for training and inference on three different games.

\section{Detailed Architecture of LogicNet}

The following code shows the architecture of LogicNet, a lightweight neural network design is proposed. Using the Traveler game as an example, the model processes images with an input resolution of 96x96, resulting in a hidden state spatial dimension of 24x24. For games such as Pong and Pac-Man, which output resolutions of 128x128, modifications to the input dimensions and corresponding network architecture are necessary to accommodate these differences.

Furthermore, considering Pong as a two-player game, the model should be adapted to receive inputs from both players. This can be achieved by introducing an additional action embedding layer and appropriately adjusting the input dimension of the linear layers to integrate dual player inputs effectively. This adjustment ensures that the network architecture remains adaptable and responsive to the specific requirements of different game dynamics.

\begin{lstlisting}[language=Python]
import torch
import torch.nn as nn

class LogicNet(nn.Module):
    def __init__(self, num_embeddings, embedding_dim=64):
        super(LogicNet, self).__init__()
        
        # Image processing: input [batch, 32, 24, 24]
        self.conv_layers = nn.Sequential(
            nn.Conv2d(32, 32, kernel_size=3, padding=1),
            nn.BatchNorm2d(32),  # Add BatchNorm to improve stability
            nn.ReLU(),
            nn.MaxPool2d(2),  # -> [batch, 32, 12, 12]
            
            nn.Conv2d(32, 64, kernel_size=3, padding=1),  # Increase number of channels
            nn.BatchNorm2d(64),
            nn.ReLU(),
            nn.MaxPool2d(2),  # -> [batch, 64, 6, 6]
        )
        
        # Integer embedding layer
        self.embedding = nn.Embedding(num_embeddings, embedding_dim)
        
        # Calculate flattened feature size: 64 * 6 * 6 = 2304
        conv_output_size = 64 * 6 * 6
        
        self.fusion_layer = nn.Sequential(
            nn.Linear(conv_output_size + embedding_dim, 512),
            nn.ReLU(),
            nn.Dropout(0.5)
        )
        
        # Final classification layer
        self.classifier = nn.Linear(512, 1)  # Binary classification
        
    def forward(self, image, action):
        # Process image
        img_features = self.conv_layers(image)
        img_features = img_features.flatten(1)  # [batch, 64 * 6 * 6]
        
        # Process integer
        action_features = self.embedding(action)
        
        # Feature fusion
        combined = torch.cat([img_features, action_features], dim=1)
        fused_features = self.fusion_layer(combined)
        
        # Classification
        predictions = self.classifier(fused_features)
        
        return predictions


if __name__ == "__main__":
    # Create model with vocabulary size of 10
    logicnet = LogicNet(10)
    
    # Create random image tensor with correct shape [batch, channels, height, width]
    image = torch.randn(1, 32, 24, 24)
    
    # Create integer tensor with a valid index (between 0 and num_embeddings-1)
    action = torch.tensor([5]).long()  # Example using index 5
    
    # Run forward pass
    output = logicnet(image, action)
    print("Output shape:", output.shape)
    print("Output value:", output)

\end{lstlisting}

\section{Details of CNN in the Spatial Module}
The CNN within the spatial module primarily comprises five convolutional layers and four downsampling max-pooling layers, which function to compress the spatial dimensions into map tokens for generation. The code is shown as follows:

\begin{lstlisting}[language=Python]
self.cnn = nn.Sequential(
    nn.Conv2d(3, 32, kernel_size=3, padding=1),
    nn.ReLU(),  
    nn.MaxPool2d(2),
    nn.Conv2d(32, 32, kernel_size=3, padding=1),
    nn.ReLU(),
    nn.MaxPool2d(2),
    nn.Conv2d(32, 32, kernel_size=3, padding=1),
    nn.ReLU(),
    nn.MaxPool2d(2),
    nn.Conv2d(32, 32, kernel_size=3, padding=1),
    nn.ReLU(),
    nn.MaxPool2d(2),
    nn.Conv2d(32, 384, kernel_size=3, padding=1),
)
\end{lstlisting}

\section{Examine the Effectiveness of Map Construction for Traveler}

\begin{figure}[h]
\centering
\includegraphics[width=1\textwidth]{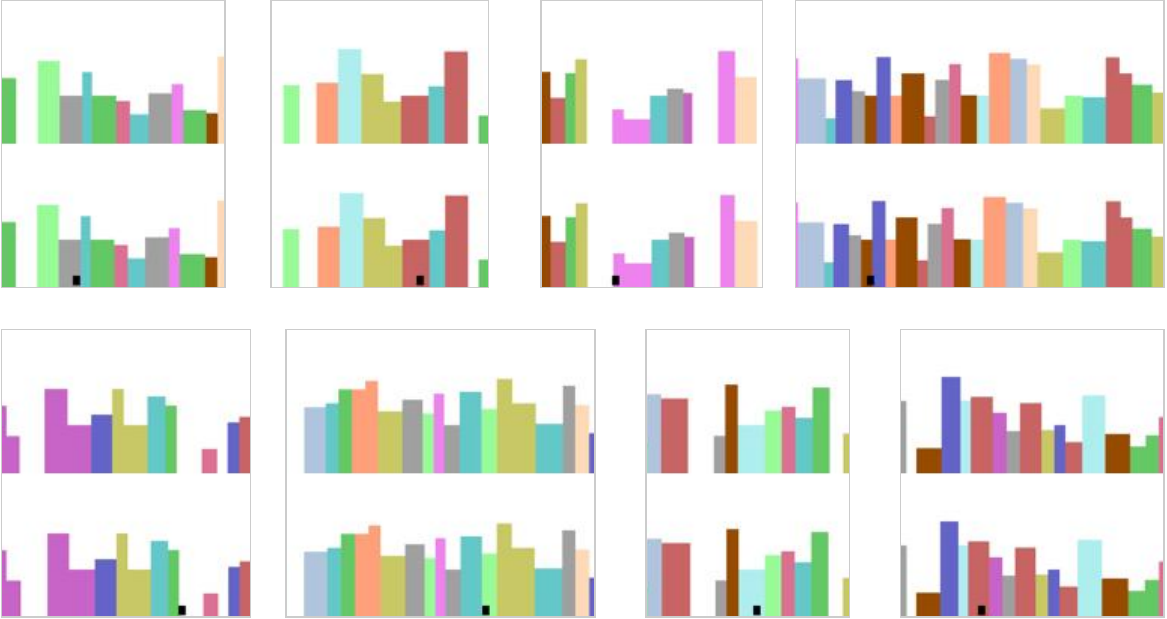}
\caption{The upper map in each pair represents the ground truth, which is directly extracted from the Pygame program, whereas the bottom map is generated using the sliding window algorithm.}
\label{fig:map1}
\end{figure}

We showcase some samples in Figure \ref{fig:map1}, where each pair represents the map of an episode in the dataset. For each pair, the upper map denotes the ground truth, which can be directly obtained from the Pygame program, while the bottom map is constructed using the sliding window algorithm. Our experiments on the evaluation dataset demonstrate the effectiveness of this method, with an average PSNR (Peak Signal-to-Noise Ratio) of 38.77. Additionally, it is worth noting that during the matching process, the black square are also included in the matching process, even though they have actually shifted to other locations. It is observed that our matching algorithm can tolerate the errors introduced by this displacement.

\section{Constructing 2D Map for Pac-Man}
\begin{figure}[h]
\centering
\includegraphics[width=0.7\textwidth]{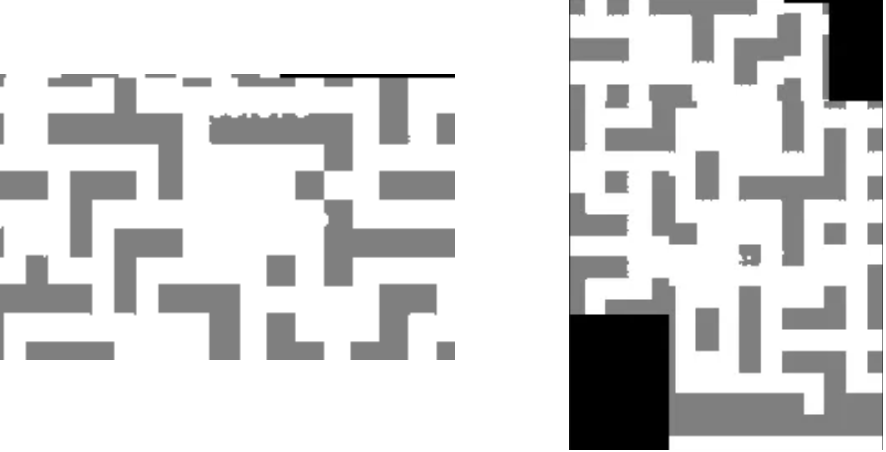}
\caption{The constructed map of two episodes in the dataset.}
\label{fig:pac2}
\end{figure}

Due to the relatively complex visuals in Pacman, we simplify the process by extracting the blue components (within a specified range) from the images to serve as features for sliding window matching. This is the result after extracting the blue components, as shown in Figure \ref{pac1}. We present the final constructed maps of two episodes in the Figure \ref{fig:pac2}. Note that since the score area in the game may overlap with the blue regions, this presents some challenges during the matching process. We leave the development of a more effective map construction algorithm for future work.

\begin{figure}[h]
\centering
\includegraphics[width=1\textwidth]{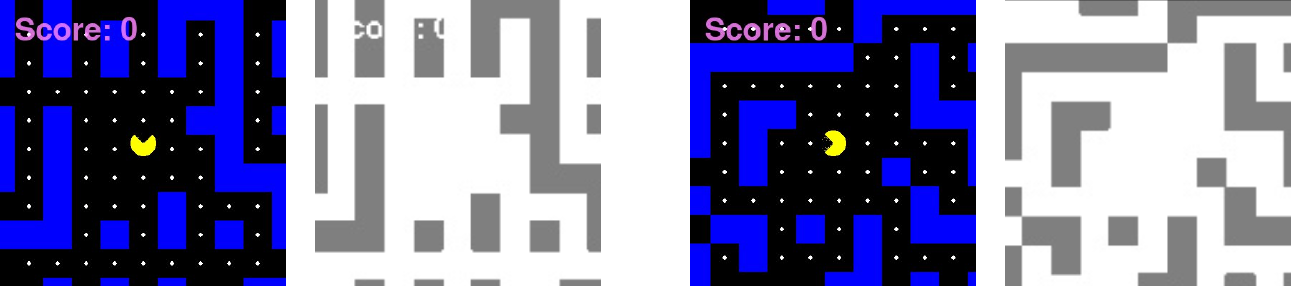}
\caption{Pre-process to extract the blue areas in the game Pac-Man for matching.}
\label{fig:pac1}
\end{figure}

\section{Details of Using LLMs for Creating Games}

The data collection process guided by LLMs is shown as follows:

\begin{itemize}
    \item \textbf{Basic Demo}: Start by describing the game we want to create using a prompt. This will help generate a simple Pygame demo of the basic game.
    
    \item \textbf{Refinement}: Adjust and improve the demo based on specific needs. This involves deciding where to place numbers on the game screen. Make sure the numbers are big enough so they're clear and can be easily used by the VAE for encoding and decoding.
    
    \item \textbf{Automated Sequences}: Use prompts to design ways for the AI to play the game, either by creating intelligent exploration strategies or by setting up a rule-based action sequence. The goal is to make sure the AI explores every possible game state.
    
    \item \textbf{Multiprocess Sampling}: Run the game in a headless mode using 96 processes to gather data fast. Typically, we can collect data from 100K game episodes in less than an hour.
\end{itemize}

\section{Details of Models Provided by PGG}

We inherit the models provided by PGG \cite{yang2024playable} and implement minor modifications, as outlined in Table \ref{tab:vae} and \ref{tab:dit}.

\begin{table}[h]
\centering
\caption{The configuration of VAE.}
\begin{tabular}{c|c}
\hline
\textbf{Item} & \textbf{Value} \\ \hline
Batch Size & 512 \\ \hline
Channels & 128 \\ \hline
Channel Multiplier & [1, 2, 4] \\ \hline
ResNet Block Number & 2 \\ \hline
KL Loss Weight & 1e-6 \\ \hline
Perceptual Loss Weight & 1 \\ \hline
Learning Rate & 4.5e-6 \\ \hline
\end{tabular}
\label{tab:vae}
\end{table}

\begin{table}[h]
\centering
\caption{The configuration of DiT.}
\begin{tabular}{c|c}
\hline
\textbf{Item} & \textbf{Value} \\ \hline
Batch Size & 512 \\ \hline
Diffusion Steps & 1000 \\ \hline
Noise Schedule & linear \\ \hline
Sequence Length & 32 \\ \hline
DiT Depth & 12 \\ \hline
DiT Hidden Size & 384 \\ \hline
DiT Patch Size & 2 \\ \hline
DiT Num Heads & 6 \\ \hline
Time Embedding Dimension & 192 \\ \hline
Action Embedding Dimension & 192 \\ \hline
Learning Rate & 3e-4 \\ \hline
\end{tabular}
\label{tab:dit}
\end{table}

\section{Detailed Architecture of Valid Action Model and Valid Numerical Model}

The architecture of the Valid Action Model is presented below. Once trained, this model processes continuous inference on two observed frames to determine whether the predicted results are consistent with the actual actions input by the player. We use the game Traveler as an example, with an input resolution of 96x96. For other games that require higher resolutions, the input dimensions can be appropriately adjusted. Accuracy is employed as the metric for evaluation.

\begin{lstlisting}[language=Python]
import torch
import torch.nn as nn

class ValidActionModel(nn.Module):
    def __init__(self):
        super(ValidActionModel, self).__init__()
        
        # CNN encoder - these weights are shared between two images
        self.encoder = nn.Sequential(
            nn.Conv2d(3, 32, kernel_size=3, stride=2, padding=1),  # 48x48x32
            nn.ReLU(),
            nn.BatchNorm2d(32),
            
            nn.Conv2d(32, 64, kernel_size=3, stride=2, padding=1),  # 24x24x64
            nn.ReLU(),
            nn.BatchNorm2d(64),
            
            nn.Conv2d(64, 128, kernel_size=3, stride=2, padding=1),  # 12x12x128
            nn.ReLU(),
            nn.BatchNorm2d(128),
            
            nn.Conv2d(128, 256, kernel_size=3, stride=2, padding=1),  # 6x6x256
            nn.ReLU(),
            nn.BatchNorm2d(256),
            
            nn.Flatten()  # 6*6*256 = 9216
        )
        
        # MLP classifier
        self.classifier = nn.Sequential(
            nn.Linear(9216 * 2, 1024),  # *2 because we have features from two images
            nn.ReLU(),
            nn.Dropout(0.5),
            
            nn.Linear(1024, 256),
            nn.ReLU(),
            nn.Dropout(0.3),
            
            nn.Linear(256, 3)  # Output 3 classes
        )
        
    def forward(self, img1, img2):
        # Encode both images
        feat1 = self.encoder(img1)  # [batch_size, 9216]
        feat2 = self.encoder(img2)  # [batch_size, 9216]
        
        # Concatenate features
        combined = torch.cat([feat1, feat2], dim=1)  # [batch_size, 9216*2]
        
        # Classification
        output = self.classifier(combined)
        
        return output


if __name__ == '__main__':
    model = ValidActionModel()
    img1 = torch.randn(1, 3, 96, 96)
    img2 = torch.randn(1, 3, 96, 96)
    output = model(img1, img2)
    print(output)
\end{lstlisting}

The architecture of the Valid Numerical Model is illustrated below. Once trained, the model performs inference by receiving a single frame input and an action to determine whether a specific event, defined as the ground truth, will occur. For our enhanced architecture, we can utilize an external score record to verify if an event has been triggered. For the baseline model, we employ a checkpoint from the TrOCR \cite{li2023trocr} printed text recognition system to obtain the current score. It has been observed that TrOCR can almost accurately recognize text within game observations. We use the game Traveler as an example, with an input resolution of 96x96. For other games that require higher resolutions, the input dimensions can be appropriately adjusted.

\begin{lstlisting}[language=Python]
import torch
import torch.nn as nn

class ValidNumericalModel(nn.Module):
    def __init__(self):
        super(ValidNumericalModel, self).__init__()
        
        # CNN encoder - single image
        self.encoder = nn.Sequential(
            nn.Conv2d(3, 32, kernel_size=3, stride=2, padding=1),  # 48x48x32
            nn.ReLU(),
            nn.BatchNorm2d(32),
            
            nn.Conv2d(32, 64, kernel_size=3, stride=2, padding=1),  # 24x24x64
            nn.ReLU(),
            nn.BatchNorm2d(64),
            
            nn.Conv2d(64, 128, kernel_size=3, stride=2, padding=1),  # 12x12x128
            nn.ReLU(),
            nn.BatchNorm2d(128),
            
            nn.Conv2d(128, 256, kernel_size=3, stride=2, padding=1),  # 6x6x256
            nn.ReLU(),
            nn.BatchNorm2d(256),
            
            nn.Flatten()  # 6*6*256 = 9216
        )
        
        # Action embedding layer
        self.action_embedding = nn.Embedding(3, 128)
        
        # MLP classifier
        self.classifier = nn.Sequential(
            nn.Linear(9216 + 128, 1024),  # Image features + action embedding
            nn.ReLU(),
            nn.Dropout(0.5),
            
            nn.Linear(1024, 256),
            nn.ReLU(),
            nn.Dropout(0.3),
            
            nn.Linear(256, 3) 
        )
        
    def forward(self, img, action):
        # Encode image
        img_feat = self.encoder(img)  # [batch_size, 9216]
        
        # Encode action
        action_feat = self.action_embedding(action.long())  # [batch_size, 128]
        
        # Concatenate features
        combined = torch.cat([img_feat, action_feat], dim=1)  # [batch_size, 9216+128]
        
        # Classification
        output = self.classifier(combined)
        
        return output


if __name__ == '__main__':
    model = ValidNumericalModel()
    img = torch.randn(1, 3, 96, 96)
    action = torch.randn(1, 3) 
    output = model(img, action)
    print(output)
\end{lstlisting}

\section{Visualizations of Failure Cases}

\begin{figure}[h]
\centering
\includegraphics[width=1\textwidth]{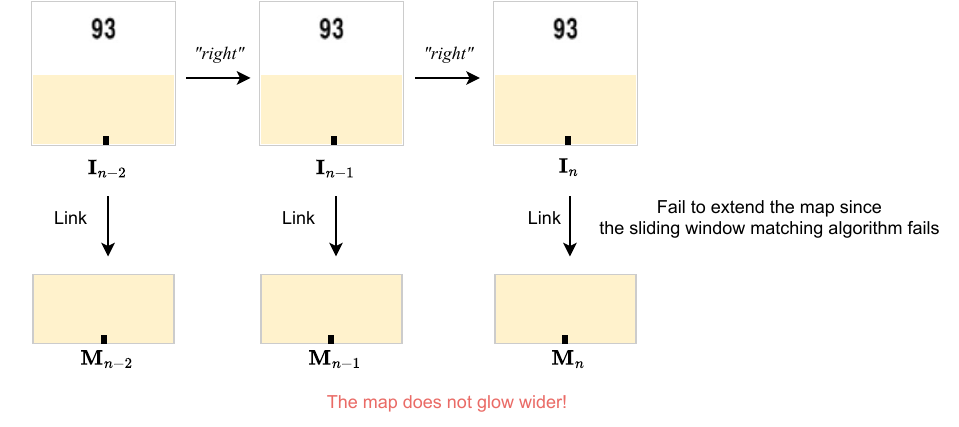}
\caption{The failure case for the sliding window matching algorithm in the case of single-color background.}
\label{fig:failure1}
\end{figure}

As demonstrated in Figure \ref{fig:failure1}, our spatial module will fail when the background comprises solely a single-colored building. Regardless of the movements made by the black square, methods based on sliding window matching fail to extend the map because the PSNR values remain high even when the window's center is kept at its original position. A promising improvement is to replace algorithm-based matching with neural network predictions. This approach requires taking into account the actions for prediction and necessitates appropriate data training to handle these corner cases effectively.

\begin{figure}[h]
\centering
\includegraphics[width=1\textwidth]{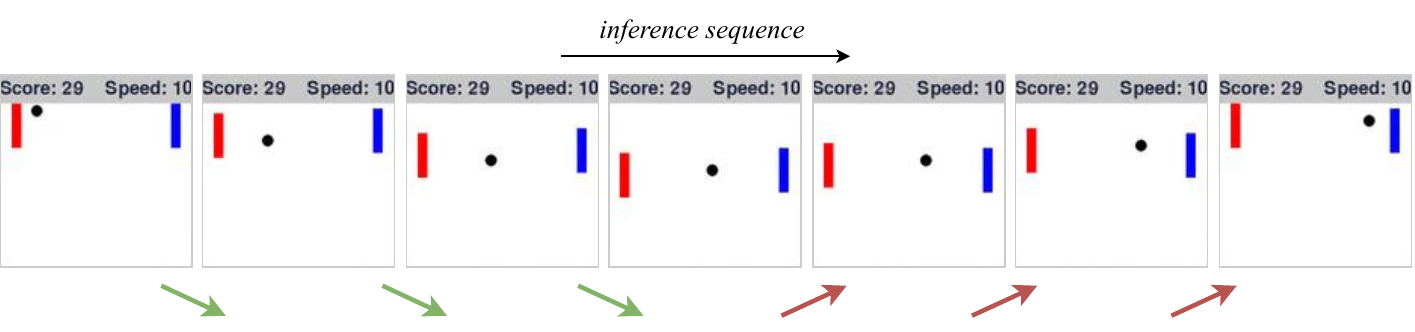}
\caption{The failure case on the physical laws. During moving down (green arrows), the black ball suddenly move up (red arrows) which is not aligned with its original trajectory.}
\label{fig:failure2}
\end{figure}

We observe that without strong supervisory features, such as physical laws, the model exhibits some instability during the inference process. As illustrated in Figure \ref{fig:failure2}, the black ball unexpectedly rises while it is supposed to be falling, which can influence the players' judgment and affect the gaming experience. We hypothesize that scaling up the model size or adopting a more robust memory mechanism, replacing the current RNN-like diffusion, may mitigate this issue.

{
    \small
    \bibliographystyle{ieeenat_fullname}
    \bibliography{main}

\begin{thebibliography}{48}
\providecommand{\natexlab}[1]{#1}
\providecommand{\url}[1]{\texttt{#1}}
\expandafter\ifx\csname urlstyle\endcsname\relax
  \providecommand{\doi}[1]{doi: #1}\else
  \providecommand{\doi}{doi: \begingroup \urlstyle{rm}\Url}\fi

\bibitem[Bruce et~al.(2024)Bruce, Dennis, Edwards, Parker-Holder, Shi, Hughes, Lai, Mavalankar, Steigerwald, Apps, et~al.]{bruce2024genie}
Jake Bruce, Michael~D Dennis, Ashley Edwards, Jack Parker-Holder, Yuge Shi, Edward Hughes, Matthew Lai, Aditi Mavalankar, Richie Steigerwald, Chris Apps, et~al.
\newblock Genie: Generative interactive environments.
\newblock In \emph{Forty-first International Conference on Machine Learning}, 2024.

\bibitem[Che et~al.(2024)Che, He, Liu, Jin, and Chen]{che2024gamegen}
Haoxuan Che, Xuanhua He, Quande Liu, Cheng Jin, and Hao Chen.
\newblock Gamegen-x: Interactive open-world game video generation.
\newblock \emph{arXiv preprint arXiv:2411.00769}, 2024.

\bibitem[Chen et~al.(2025)Chen, Mart{\'\i}~Mons{\'o}, Du, Simchowitz, Tedrake, and Sitzmann]{chen2025diffusion}
Boyuan Chen, Diego Mart{\'\i}~Mons{\'o}, Yilun Du, Max Simchowitz, Russ Tedrake, and Vincent Sitzmann.
\newblock Diffusion forcing: Next-token prediction meets full-sequence diffusion.
\newblock \emph{Advances in Neural Information Processing Systems}, 37:\penalty0 24081--24125, 2025.

\bibitem[Chen et~al.(2023{\natexlab{a}})Chen, Huang, Lv, Cui, Chen, and Wei]{chen2023textdiffuser}
Jingye Chen, Yupan Huang, Tengchao Lv, Lei Cui, Qifeng Chen, and Furu Wei.
\newblock Textdiffuser: Diffusion models as text painters.
\newblock \emph{Advances in Neural Information Processing Systems}, 36:\penalty0 9353--9387, 2023{\natexlab{a}}.

\bibitem[Chen et~al.(2024)Chen, Huang, Lv, Cui, Chen, and Wei]{chen2024textdiffuser}
Jingye Chen, Yupan Huang, Tengchao Lv, Lei Cui, Qifeng Chen, and Furu Wei.
\newblock Textdiffuser-2: Unleashing the power of language models for text rendering.
\newblock In \emph{European Conference on Computer Vision}, pages 386--402. Springer, 2024.

\bibitem[Chen et~al.(2023{\natexlab{b}})Chen, Ji, Wu, Wu, Xie, Li, Xia, Xiao, and Lin]{chen2023control}
Weifeng Chen, Yatai Ji, Jie Wu, Hefeng Wu, Pan Xie, Jiashi Li, Xin Xia, Xuefeng Xiao, and Liang Lin.
\newblock Control-a-video: Controllable text-to-video generation with diffusion models.
\newblock \emph{arXiv e-prints}, pages arXiv--2305, 2023{\natexlab{b}}.

\bibitem[claude(2024)]{claude}
claude.
\newblock Link: https://www.anthropic.com/news/claude-3-5-sonnet, 2024.

\bibitem[{Epic Games}()]{unrealengine}
{Epic Games}.
\newblock Unreal engine.

\bibitem[Esser et~al.(2024)Esser, Kulal, Blattmann, Entezari, M{\"u}ller, Saini, Levi, Lorenz, Sauer, Boesel, et~al.]{esser2024scaling}
Patrick Esser, Sumith Kulal, Andreas Blattmann, Rahim Entezari, Jonas M{\"u}ller, Harry Saini, Yam Levi, Dominik Lorenz, Axel Sauer, Frederic Boesel, et~al.
\newblock Scaling rectified flow transformers for high-resolution image synthesis.
\newblock In \emph{Forty-first international conference on machine learning}, 2024.

\bibitem[Feng et~al.(2024)Feng, Zhang, Yang, Xiao, Shu, Liu, Zheng, Huang, Liu, and Zhang]{feng2024matrix}
Ruili Feng, Han Zhang, Zhantao Yang, Jie Xiao, Zhilei Shu, Zhiheng Liu, Andy Zheng, Yukun Huang, Yu Liu, and Hongyang Zhang.
\newblock The matrix: Infinite-horizon world generation with real-time moving control.
\newblock \emph{arXiv preprint arXiv:2412.03568}, 2024.

\bibitem[flux(2024)]{flux}
flux.
\newblock Link: https://github.com/black-forest-labs/flux, 2024.

\bibitem[genie 2(2024)]{genie-2}
genie 2.
\newblock Link: https://deepmind.google/discover/blog/genie-2-a-large-scale-foundation-world-model/, 2024.

\bibitem[Haas(2014)]{unity}
John~K Haas.
\newblock A history of the unity game engine.
\newblock 2014.

\bibitem[He et~al.(2024{\natexlab{a}})He, Xu, Guo, Wetzstein, Dai, Li, and Yang]{he2024cameractrl}
Hao He, Yinghao Xu, Yuwei Guo, Gordon Wetzstein, Bo Dai, Hongsheng Li, and Ceyuan Yang.
\newblock Cameractrl: Enabling camera control for text-to-video generation.
\newblock \emph{arXiv preprint arXiv:2404.02101}, 2024{\natexlab{a}}.

\bibitem[He et~al.(2024{\natexlab{b}})He, Liu, Chen, Tian, Liu, Chi, Liu, Yuan, Xing, Wang, et~al.]{he2024llms}
Yingqing He, Zhaoyang Liu, Jingye Chen, Zeyue Tian, Hongyu Liu, Xiaowei Chi, Runtao Liu, Ruibin Yuan, Yazhou Xing, Wenhai Wang, et~al.
\newblock Llms meet multimodal generation and editing: A survey.
\newblock \emph{arXiv preprint arXiv:2405.19334}, 2024{\natexlab{b}}.

\bibitem[Heusel et~al.(2017)Heusel, Ramsauer, Unterthiner, Nessler, and Hochreiter]{fid}
Martin Heusel, Hubert Ramsauer, Thomas Unterthiner, Bernhard Nessler, and Sepp Hochreiter.
\newblock Gans trained by a two time-scale update rule converge to a local nash equilibrium.
\newblock \emph{Advances in neural information processing systems}, 30, 2017.

\bibitem[Ho et~al.(2020)Ho, Jain, and Abbeel]{ho2020denoising}
Jonathan Ho, Ajay Jain, and Pieter Abbeel.
\newblock Denoising diffusion probabilistic models.
\newblock \emph{Advances in neural information processing systems}, 33:\penalty0 6840--6851, 2020.

\bibitem[Hu and Xu(2023)]{hu2023videocontrolnet}
Zhihao Hu and Dong Xu.
\newblock Videocontrolnet: A motion-guided video-to-video translation framework by using diffusion model with controlnet.
\newblock \emph{arXiv preprint arXiv:2307.14073}, 2023.

\bibitem[Huang et~al.(2023)Huang, Chen, Liu, Shen, Zhao, and Zhou]{huang2023composer}
Lianghua Huang, Di Chen, Yu Liu, Yujun Shen, Deli Zhao, and Jingren Zhou.
\newblock Composer: Creative and controllable image synthesis with composable conditions.
\newblock \emph{arXiv preprint arXiv:2302.09778}, 2023.

\bibitem[Kanervisto et~al.(2025)Kanervisto, Bignell, Wen, Grayson, Georgescu, Valcarcel~Macua, Tan, Rashid, Pearce, Cao, et~al.]{kanervisto2025world}
Anssi Kanervisto, Dave Bignell, Linda~Yilin Wen, Martin Grayson, Raluca Georgescu, Sergio Valcarcel~Macua, Shan~Zheng Tan, Tabish Rashid, Tim Pearce, Yuhan Cao, et~al.
\newblock World and human action models towards gameplay ideation.
\newblock \emph{Nature}, 638\penalty0 (8051):\penalty0 656--663, 2025.

\bibitem[Kim et~al.(2020)Kim, Zhou, Philion, Torralba, and Fidler]{kim2020learning}
Seung~Wook Kim, Yuhao Zhou, Jonah Philion, Antonio Torralba, and Sanja Fidler.
\newblock Learning to simulate dynamic environments with gamegan.
\newblock In \emph{Proceedings of the IEEE/CVF Conference on Computer Vision and Pattern Recognition}, pages 1231--1240, 2020.

\bibitem[Kingma and Welling(2013)]{kingma2013auto}
Diederik~P Kingma and Max Welling.
\newblock Auto-encoding variational bayes.
\newblock \emph{arXiv preprint arXiv:1312.6114}, 2013.

\bibitem[Kondratyuk et~al.(2023)Kondratyuk, Yu, Gu, Lezama, Huang, Schindler, Hornung, Birodkar, Yan, Chiu, et~al.]{kondratyuk2023videopoet}
Dan Kondratyuk, Lijun Yu, Xiuye Gu, Jos{\'e} Lezama, Jonathan Huang, Grant Schindler, Rachel Hornung, Vighnesh Birodkar, Jimmy Yan, Ming-Chang Chiu, et~al.
\newblock Videopoet: A large language model for zero-shot video generation.
\newblock \emph{arXiv preprint arXiv:2312.14125}, 2023.

\bibitem[Li et~al.(2024{\natexlab{a}})Li, Li, Wadhwa, Pritch, Jacobs, Rubinstein, Bansal, and Ruiz]{li2024unbounded}
Jialu Li, Yuanzhen Li, Neal Wadhwa, Yael Pritch, David~E Jacobs, Michael Rubinstein, Mohit Bansal, and Nataniel Ruiz.
\newblock Unbounded: A generative infinite game of character life simulation.
\newblock \emph{arXiv preprint arXiv:2410.18975}, 2024{\natexlab{a}}.

\bibitem[Li et~al.(2023)Li, Lv, Chen, Cui, Lu, Florencio, Zhang, Li, and Wei]{li2023trocr}
Minghao Li, Tengchao Lv, Jingye Chen, Lei Cui, Yijuan Lu, Dinei Florencio, Cha Zhang, Zhoujun Li, and Furu Wei.
\newblock Trocr: Transformer-based optical character recognition with pre-trained models.
\newblock In \emph{Proceedings of the AAAI conference on artificial intelligence}, pages 13094--13102, 2023.

\bibitem[Li et~al.(2024{\natexlab{b}})Li, Yang, Kuang, Wu, Wang, Xiao, and Chen]{li2024controlnet++}
Ming Li, Taojiannan Yang, Huafeng Kuang, Jie Wu, Zhaoning Wang, Xuefeng Xiao, and Chen Chen.
\newblock Controlnet++: Improving conditional controls with efficient consistency feedback: Project page: liming-ai. github. io/controlnet\_plus\_plus.
\newblock In \emph{European Conference on Computer Vision}, pages 129--147. Springer, 2024{\natexlab{b}}.

\bibitem[Menapace et~al.(2021)Menapace, Lathuiliere, Tulyakov, Siarohin, and Ricci]{menapace2021playable}
Willi Menapace, Stephane Lathuiliere, Sergey Tulyakov, Aliaksandr Siarohin, and Elisa Ricci.
\newblock Playable video generation.
\newblock In \emph{Proceedings of the IEEE/CVF Conference on Computer Vision and Pattern Recognition}, pages 10061--10070, 2021.

\bibitem[Menapace et~al.(2024)Menapace, Siarohin, Lathuili{\`e}re, Achlioptas, Golyanik, Tulyakov, and Ricci]{menapace2024promptable}
Willi Menapace, Aliaksandr Siarohin, St{\'e}phane Lathuili{\`e}re, Panos Achlioptas, Vladislav Golyanik, Sergey Tulyakov, and Elisa Ricci.
\newblock Promptable game models: Text-guided game simulation via masked diffusion models.
\newblock \emph{ACM Transactions on Graphics}, 43\penalty0 (2):\penalty0 1--16, 2024.

\bibitem[Mou et~al.(2024)Mou, Wang, Xie, Wu, Zhang, Qi, and Shan]{mou2024t2i}
Chong Mou, Xintao Wang, Liangbin Xie, Yanze Wu, Jian Zhang, Zhongang Qi, and Ying Shan.
\newblock T2i-adapter: Learning adapters to dig out more controllable ability for text-to-image diffusion models.
\newblock In \emph{Proceedings of the AAAI conference on artificial intelligence}, pages 4296--4304, 2024.

\bibitem[oasis(2024)]{oasis}
oasis.
\newblock Link: https://github.com/etched-ai/open-oasis, 2024.

\bibitem[Ouyang et~al.(2024)Ouyang, Wang, Xiao, Bai, Zhang, Zheng, Zhou, Chen, and Shen]{ouyang2024codef}
Hao Ouyang, Qiuyu Wang, Yuxi Xiao, Qingyan Bai, Juntao Zhang, Kecheng Zheng, Xiaowei Zhou, Qifeng Chen, and Yujun Shen.
\newblock Codef: Content deformation fields for temporally consistent video processing.
\newblock In \emph{Proceedings of the IEEE/CVF Conference on Computer Vision and Pattern Recognition}, pages 8089--8099, 2024.

\bibitem[Peebles and Xie(2023)]{peebles2023scalable}
William Peebles and Saining Xie.
\newblock Scalable diffusion models with transformers.
\newblock In \emph{Proceedings of the IEEE/CVF international conference on computer vision}, pages 4195--4205, 2023.

\bibitem[Peng et~al.(2024)Peng, Wang, Zhang, Li, Yang, and Jia]{peng2024controlnext}
Bohao Peng, Jian Wang, Yuechen Zhang, Wenbo Li, Ming-Chang Yang, and Jiaya Jia.
\newblock Controlnext: Powerful and efficient control for image and video generation.
\newblock \emph{arXiv preprint arXiv:2408.06070}, 2024.

\bibitem[Rombach et~al.(2022)Rombach, Blattmann, Lorenz, Esser, and Ommer]{rombach2022high}
Robin Rombach, Andreas Blattmann, Dominik Lorenz, Patrick Esser, and Bj{\"o}rn Ommer.
\newblock High-resolution image synthesis with latent diffusion models.
\newblock In \emph{Proceedings of the IEEE/CVF conference on computer vision and pattern recognition}, pages 10684--10695, 2022.

\bibitem[Sudhakaran et~al.(2023)Sudhakaran, Gonz{\'a}lez-Duque, Freiberger, Glanois, Najarro, and Risi]{sudhakaran2023mariogpt}
Shyam Sudhakaran, Miguel Gonz{\'a}lez-Duque, Matthias Freiberger, Claire Glanois, Elias Najarro, and Sebastian Risi.
\newblock Mariogpt: Open-ended text2level generation through large language models.
\newblock \emph{Advances in Neural Information Processing Systems}, 36:\penalty0 54213--54227, 2023.

\bibitem[Unterthiner et~al.(2018)Unterthiner, Van~Steenkiste, Kurach, Marinier, Michalski, and Gelly]{fvd}
Thomas Unterthiner, Sjoerd Van~Steenkiste, Karol Kurach, Raphael Marinier, Marcin Michalski, and Sylvain Gelly.
\newblock Towards accurate generative models of video: A new metric \& challenges.
\newblock \emph{arXiv preprint arXiv:1812.01717}, 2018.

\bibitem[Valevski et~al.(2024)Valevski, Leviathan, Arar, and Fruchter]{valevski2024diffusion}
Dani Valevski, Yaniv Leviathan, Moab Arar, and Shlomi Fruchter.
\newblock Diffusion models are real-time game engines.
\newblock \emph{arXiv preprint arXiv:2408.14837}, 2024.

\bibitem[Yang et~al.(2024{\natexlab{a}})Yang, Li, Fang, Chen, Yu, Fu, Yang, and Ye]{yang2024playable}
Mingyu Yang, Junyou Li, Zhongbin Fang, Sheng Chen, Yangbin Yu, Qiang Fu, Wei Yang, and Deheng Ye.
\newblock Playable game generation.
\newblock \emph{arXiv preprint arXiv:2412.00887}, 2024{\natexlab{a}}.

\bibitem[Yang et~al.(2024{\natexlab{b}})Yang, Hou, Huang, Ma, Wan, Zhang, Chen, and Liao]{yang2024direct}
Shiyuan Yang, Liang Hou, Haibin Huang, Chongyang Ma, Pengfei Wan, Di Zhang, Xiaodong Chen, and Jing Liao.
\newblock Direct-a-video: Customized video generation with user-directed camera movement and object motion.
\newblock In \emph{ACM SIGGRAPH 2024 Conference Papers}, pages 1--12, 2024{\natexlab{b}}.

\bibitem[ying(2024)]{ying}
ying.
\newblock Link: https://giantailab.github.io/yinggame/, 2024.

\bibitem[Yu et~al.(2025)Yu, Qin, Wang, Wan, Zhang, and Liu]{yu2025gamefactory}
Jiwen Yu, Yiran Qin, Xintao Wang, Pengfei Wan, Di Zhang, and Xihui Liu.
\newblock Gamefactory: Creating new games with generative interactive videos.
\newblock \emph{arXiv preprint arXiv:2501.08325}, 2025.

\bibitem[Zhang et~al.(2023{\natexlab{a}})Zhang, Rao, and Agrawala]{zhang2023adding}
Lvmin Zhang, Anyi Rao, and Maneesh Agrawala.
\newblock Adding conditional control to text-to-image diffusion models.
\newblock In \emph{Proceedings of the IEEE/CVF international conference on computer vision}, pages 3836--3847, 2023{\natexlab{a}}.

\bibitem[Zhang et~al.(2024{\natexlab{a}})Zhang, Zhai, Bautista, Miao, Toshev, Susskind, and Gu]{zhang2024world}
Qihang Zhang, Shuangfei Zhai, Miguel~Angel Bautista, Kevin Miao, Alexander Toshev, Joshua Susskind, and Jiatao Gu.
\newblock World-consistent video diffusion with explicit 3d modeling.
\newblock \emph{arXiv preprint arXiv:2412.01821}, 2024{\natexlab{a}}.

\bibitem[Zhang et~al.(2018)Zhang, Isola, Efros, Shechtman, and Wang]{zhang2018unreasonable}
Richard Zhang, Phillip Isola, Alexei~A Efros, Eli Shechtman, and Oliver Wang.
\newblock The unreasonable effectiveness of deep features as a perceptual metric.
\newblock In \emph{Proceedings of the IEEE conference on computer vision and pattern recognition}, pages 586--595, 2018.

\bibitem[Zhang et~al.(2023{\natexlab{b}})Zhang, Wei, Jiang, Zhang, Zuo, and Tian]{zhang2023controlvideo}
Yabo Zhang, Yuxiang Wei, Dongsheng Jiang, Xiaopeng Zhang, Wangmeng Zuo, and Qi Tian.
\newblock Controlvideo: Training-free controllable text-to-video generation.
\newblock \emph{arXiv preprint arXiv:2305.13077}, 2023{\natexlab{b}}.

\bibitem[Zhang et~al.(2024{\natexlab{b}})Zhang, Kang, Zhang, Ding, Zhao, and Yue]{zhang2024interactivevideo}
Yiyuan Zhang, Yuhao Kang, Zhixin Zhang, Xiaohan Ding, Sanyuan Zhao, and Xiangyu Yue.
\newblock Interactivevideo: User-centric controllable video generation with synergistic multimodal instructions.
\newblock \emph{arXiv preprint arXiv:2402.03040}, 2024{\natexlab{b}}.

\bibitem[Zhao et~al.(2023)Zhao, Chen, Chen, Bao, Hao, Yuan, and Wong]{zhao2023uni}
Shihao Zhao, Dongdong Chen, Yen-Chun Chen, Jianmin Bao, Shaozhe Hao, Lu Yuan, and Kwan-Yee~K Wong.
\newblock Uni-controlnet: All-in-one control to text-to-image diffusion models.
\newblock \emph{Advances in Neural Information Processing Systems}, 36:\penalty0 11127--11150, 2023.

\bibitem[Zhou et~al.(2024)Zhou, Wang, Nie, Lin, Yu, Yu, and Wang]{zhou2024trackgo}
Haitao Zhou, Chuang Wang, Rui Nie, Jinxiao Lin, Dongdong Yu, Qian Yu, and Changhu Wang.
\newblock Trackgo: A flexible and efficient method for controllable video generation.
\newblock \emph{arXiv preprint arXiv:2408.11475}, 2024.

\end{thebibliography}
}

\end{document}